\documentclass[runningheads]{llncs}
\usepackage{graphicx}
\usepackage{tikz}
\usepackage{comment}
\usepackage{amsmath,amssymb} % define this before the line numbering.
\usepackage{color}
\usepackage{multirow}
\usepackage{makecell}
\usepackage{booktabs}
\usepackage{algorithm}
\usepackage{algorithmic}

\usepackage[accsupp]{axessibility}  % Improves PDF readability for those with disabilities.

\usepackage[width=122mm,left=12mm,paperwidth=146mm,height=193mm,top=12mm,paperheight=217mm]{geometry}

\begin{document}
% \linenumbers
\pagestyle{headings}
\mainmatter
\def\ECCVSubNumber{2708}  % Insert your submission number here

\author{Dong Jing\quad Shuo Zhang \quad Song Chang\quad Youfang Lin }
\institute{Beijing Jiaotong University
}

\title{Light Field Raindrop Removal via 4D Re-sampling} % Replace with your title

%\begin{comment}
% \titlerunning{ECCV-22 submission ID \ECCVSubNumber} 
% \authorrunning{ECCV-22 submission ID \ECCVSubNumber} 
% \author{Anonymous ECCV submission}
% \institute{Paper ID \ECCVSubNumber}
%\end{comment}
%******************

% \begin{comment}
% % \titlerunning{Abbreviated paper title}
% % If the paper title is too long for the running head, you can set
% % an abbreviated paper title here
% \end{comment}
%******************
\maketitle

\begin{abstract}
% Raindrop attached to a glass window or camera lens can severely hamper the visibility of background scene and considerably degrade an image.
% The goal of the raindrop removal task is to restore the background area obscured by raindrops in an image.

% Single image raindrop removal is a challenging problem due to the lack of texture details in the occlusion area.
% In contrast, the Light Field (LF) provides more abundant information by regularly and densely sampling the scene.
% We introduce the LF into the raindrop removal task to improve the performance.

The Light Field Raindrop Removal (LFRR) aims to restore the background areas obscured by raindrops in the Light Field (LF).
Compared with single image, the LF provides more abundant information by regularly and densely sampling the scene.
% The LF consists of multiple scene views captured by regular and dense sampling in the camera plane.
Since raindrops have larger disparities than the background in the LF, the majority of texture details occluded by raindrops are visible in other views.
In this paper, we propose a novel LFRR network by directly utilizing the complementary pixel information of raindrop-free areas in the input raindrop LF, which consists of the re-sampling module and the refinement module. 
Specifically, the re-sampling module generates a new LF which is less polluted by raindrops through re-sampling position predictions and the proposed 4D interpolation.
The refinement module improves the restoration of the completely occluded background areas and corrects the pixel error caused by 4D interpolation.
% Specifically, the network preliminarily restore the occluded background areas by 4D interpolation re-sampling on the raindrop LF.
% The re-sampling strategy leverages the useful pixel information in raindrop LF and effectively reduces the dependence on implicit residual prediction.
Furthermore, we carefully build the first real scene LFRR dataset for model training and validation.
Experiments demonstrate that the proposed method can effectively remove raindrops and achieves state-of-the-art performance in both background restoration and view consistency maintenance.
% Besides, we public the first LFRR real scene dataset.

\end{abstract}

\section{Introduction}

Image raindrop removal refers to the restoration of background areas obscured by raindrops in the image, which is beneficial for many high-level computer vision applications, such as object detection \cite{dai2017deformable,zhu2019deformable} and auto-driving \cite{wang2019survey}.
Most existing raindrop removal approaches are based on 2D single image \cite{quan2021removing,jiang2020multi,fu2017removing,yasarla2019uncertainty,wang2019spatial}.
However, due to the lack of texture details in the occlusion area, 2D approaches excessively depend on the surrounding pixels or the inpainting patterns learned from the training set when restoring the background, which leads to unnatural and unreliable results.
%  the severe image degradation caused by large raindrop spots and
% Therefore, single image based approaches depends more on the inpainting patterns learned from the training set when restoring the background, which leads to unnatural and unreal results.

\begin{figure}[!t]
  \centering
  \includegraphics[width=0.9\linewidth]{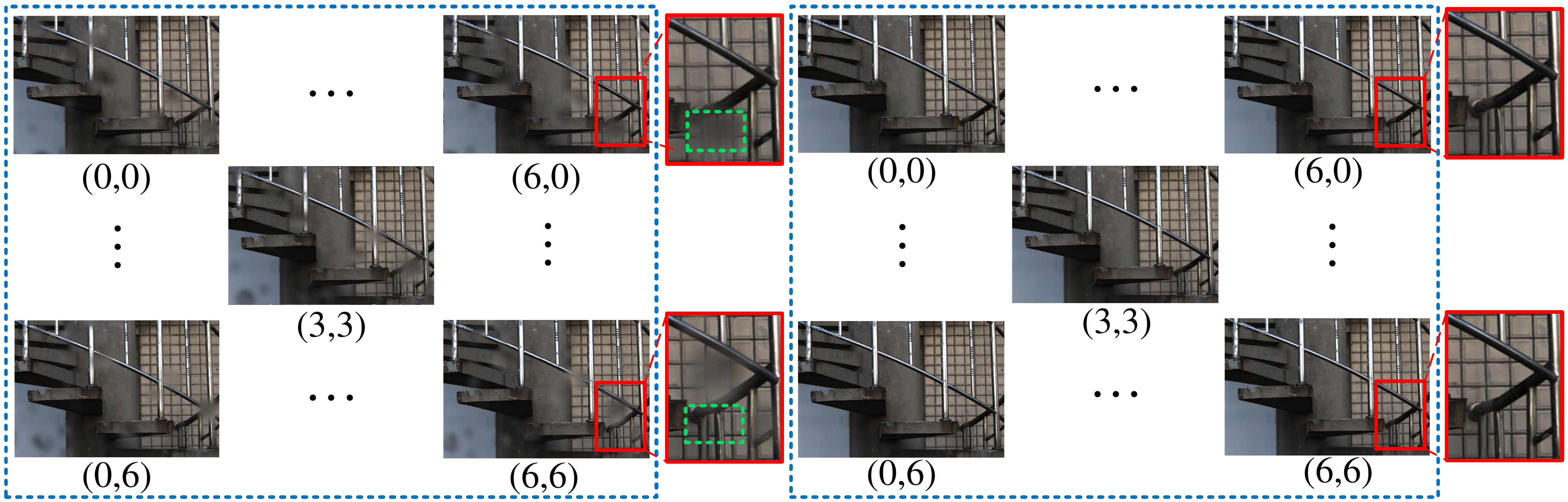}
  \caption{Left: the raindrop LF.
  Right: raindrop removal result generated by our model.
  Since raindrops are closer to the camera than backgrounds, raindrops have larger disparities.
  As shown in the left instance, the obscured texture details in green box of view$(6,0)$ are visible in the corresponding areas of view$(6,6)$.
  }
  \label{fig:intro}
\end{figure}

The Light Field (LF) has lately received great attention due to its capability of providing abundant complementary information and implicit scene depth information.
The LF consists of multiple scene views generated by regularly and densely sampling in the camera plane.
Compared with the 2D image, the LF owns two angular dimensions together with two spatial dimensions, so the LF can also be described as the 4D sampling of the scene.
In the LF, the distance a pixel moves between adjacent views is called the pixel's disparity.
% In LFs, the closer the object is to the camera, the larger the disparity it has.
Since raindrops are closer to the camera lens than the background, raindrops have larger disparities than the background in the LF.
Therefore, as shown in the left instance of Fig.\ref{fig:intro}, most texture details of the background areas obscured by raindrops are visible in other views.
In addition to the complementary information between views, structural characteristics of the LF cause raindrop removal targets of different views are highly similar and correlated, which is beneficial for learning objective designing.
Thus, we introduce the LF into the raindrop removal task to go beyond the limitation of single image.

% When taking photos on rainy days or through a glass window, raindrops adhered to camera lens or the glass window degrade the captured image quality.
% Raindrop removal refers to the restoration of the background area obscured by raindrops in images, which is beneficial for many high-level computer vision applications, such as object detection \cite{hu2018squeeze} and auto-driving \cite{wang2019survey}.
% Benefiting for open source datasets and deep learning development, single image deraining field has accumulated rich research achievements. 
% However, single image raindrop removal is still a challenging problem due to the severe image degradation caused by large raindrop spots occlusion and the lack of texture details in the occlusion area.
% 2D methods recover the raindrop area mainly through the learned inpainting patterns from the training data.

In this paper, we propose a novel convolutional neural network to address the Light Field Raindrop Removal (LFRR) problem.
% by directly utilizing the complementary pixel information in the input raindrop LF.
On one hand, we directly select raindrop-free areas from other views to restore the polluted areas by re-sampling the input raindrop LF.
On the other hand, we utilize the convolution module to implicitly refine the re-sampled LF.
% Specifically, two critical stages in our model are re-sampling and refinement.
In the re-sampling stage, the model selects pixels from the raindrop LF and generates a novel LF which has less raindrop pollution through re-sampling position prediction and the proposed 4D interpolation.
The re-sampling strategy not only leverages the complementary pixel information between different views, but also effectively reduces the dependence on implicit residual prediction.
In the refinement stage, the model generates a residual map for improving the restoration of the completely occluded areas and correcting the error caused by 4D interpolation in the re-sampled LF.
Considering that the areas with large differences between the raindrop LF and the re-sampled LF are consistent with the areas where need refinement, we devise a novel difference guided encode block adopted in the refinement module to leverage the guidance of these differences.

Furthermore, we carefully build the first real scene LFRR dataset which contains a variety of indoor and outdoor scenes to meet the need of model training and validation.
% We compare our approach with six single image deraining methods and four adjusted LF super resolution methods.
Experimental results demonstrate that our approach can effectively remove raindrops in the LF and achieves the state-of-the-art performance in both background restoration and view consistency maintenance.

In summary, our main contributions are as follows:
\begin{enumerate}
    \item In order to go beyond the limitation of single image, we introduce the LF into the raindrop removal field, which provides more abundant scene information and enhances the raindrop removal performance.
    \item We propose a novel LFRR network by re-sampling the raindrop LF and refining the re-sampled LF.
    We design a novel difference guided encode block for more effective feature encoding during the refinement.
    \item We build the first real scene LFRR dataset, including various indoor and outdoor scenes, to meet the needs of model training and validation.
\end{enumerate}

\section{Related Work}
% 这里计划写 1.去雨滴的进展； 2.光场整体处理相关的网络。
% 去雨这里通常会采用short cut + residual的方式。
% 去雨的过程会被分为雨滴的检测 + 雨滴的去除两部分。
% 去雨的改进集中在几个方面：更符合待去除目标特性的建模、多尺度、迭代的方式、数据集增强等。
% 光场这里想写一下光场end-to-end处理的一些baseline以及进展，主要是特征提取方面的，或者光场一致性的约束——EPIloss，或者MDFN这里的动态卷积。
In this section, we briefly summary the related single image raindrop removal methods and LF processing methods.
\subsection{Single Image Raindrop Removal}
% Single image deraining task aim to recover the backgrounds obscured by rain steaks, raindrops and rain haze.
% \cite{fu2017removing} first applied the residual network to deraining task. 
% They explored the impact of different learning objectives on the model performance, such as mapping directly from rain input to rain-free output, or predicting the residual map.
% They determined the learning objective of the model as the residual map between the rain image and rain-free image which is also adopted by most subsequent deraining algorithms.
Due to the lack of texture details in occluded areas, the single image raindrop removal problem is much more challenging.
The progress of this task has been stagnated until the rapid progress of deep learning in recent years.

\cite{eigen2013restoring} used three convolutional layers to remove raindrops and generated outputs with relatively poor performance.
\cite{qian2018attentive} proposed a generative adversarial network for raindrop removal, where the generative network produces the attention map via an attentive-recurrent network and applies this raindrop attention mask to generate a raindrop-free image through a contextual auto-encoder.
\cite{quan2019deep} proposed a network, in which the shape-driven attention module exploits the physical shape properties of raindrops, including closedness and roundness, and the channel attention refines the features relevant to the background layer or the raindrop layer.
\cite{shao2021uncertainty} designed a uncertainty guided multi-scale attention network which benefits from the exploration of the blur-level of raindrops, and utilized several basic modules to explore the inherent correlations of the similar raindrop patterns across scales.

\subsection{Light Field Processing}
\label{LFP}
Since there are no reference algorithms in LFRR, we briefly present some approaches in related LF based fields in this subsection.

The LF is a 4D data with unique structural characteristics.
A hot and general research topic is to design appropriate modules to extract effective information from the LF.
\cite{yeung2018light} proposed a spatial-angular separable convolution module, in which two convolution layers are separately and serially extract features in spatial or angular dimensions.
\cite{wang2020spatial} designed an interaction mechanism to incorporate decoupled spatial and angular information.
The angular feature in this model is extracted by a no-padding convolution layer which contains the global angular information.
\cite{sun2020multi} devised a module named Multi-Dimension  Fusion  Block (MDFB) which utilizes four parallel convolution layers to extract features from sub-aperture images, micro-lens images, horizontal and vertical epipolar images (EPIs) respectively.
The MDFB is also adopted in our model for LF feature extraction.

The LF spatial super-resolution (LFSSR) task has many similarities with LFRR, and LFSSR approaches are conveniently adjusted to solve the LFRR task.
Firstly, both LFRR and LFSSR process all views of the LF, while many typical LF tasks only deal with the central view of the LF, such as saliency detection \cite{jing2021occlusion}, depth estimation \cite{chen2021attention} and occlusion removal \cite{Shen2021Deocc}.
Secondly, they both are low-level image restoration tasks and aim to enhance the quality of the input LF.
Thirdly, they both need effective complementary information and structural characteristics existed within and between views of the LF.
Besides, existing LFSSR approaches up-sample features and LFs by spatial interpolation, transposed convolution or pixel shuffle.
By removing up-sampling operations, most LFSSR approaches can be utilized to handle the LFRR task.

% Most LF super resolution methods adopt the same up-sampling strategy.
% They extract features from the input LF and up-sample the features by a transposed convolution layer at the last module.
% The input LF is also up-sampled by 2D spatial interpolation or a transposed convolution layer for summation.
% Therefore, these methods are easy to adjust to the LFRR by removing up-sampling operations.

\begin{figure}[!h]
  \centering
  \includegraphics[width=0.9\linewidth]{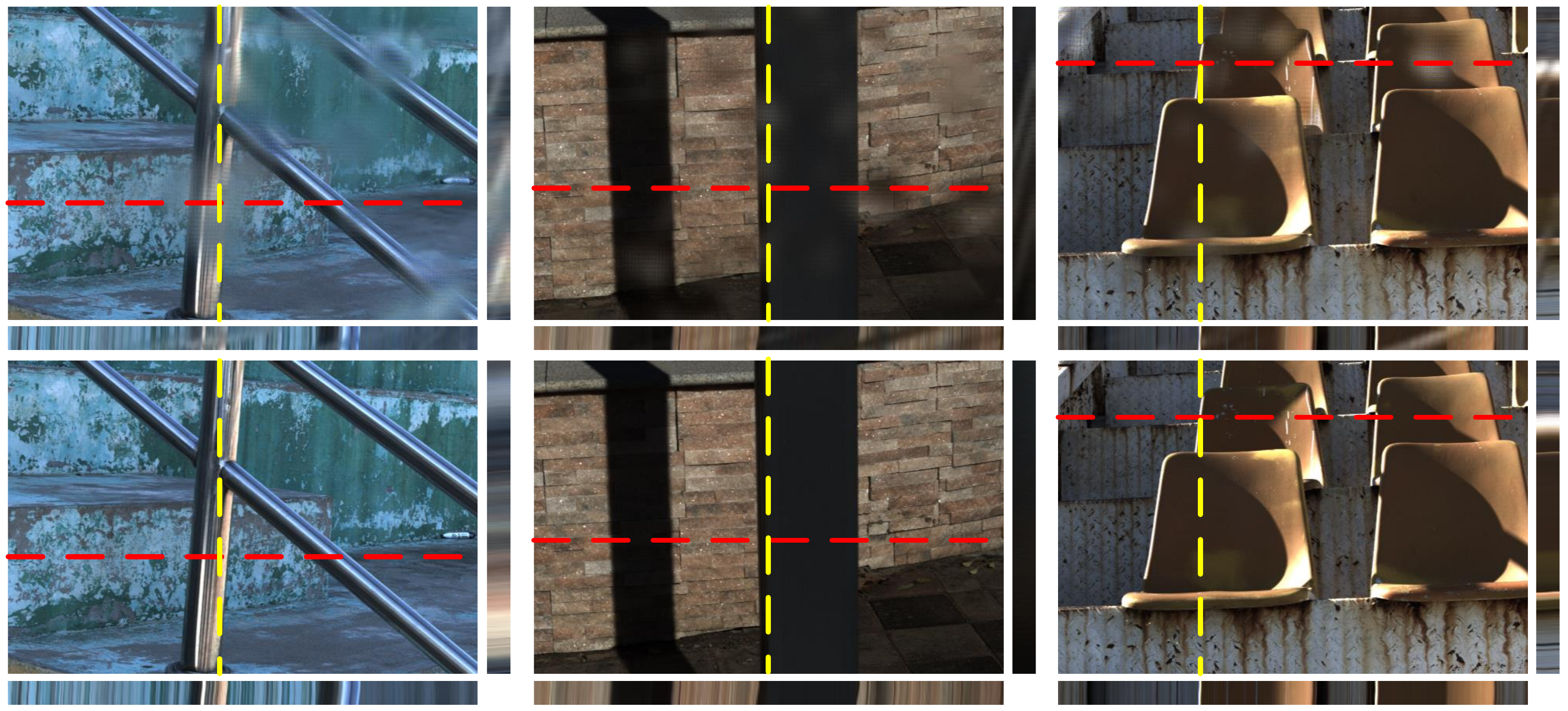}
  \caption{
  Samples of our dataset.
  Top: the central views of raindrop LFs.
  Bottom: the corresponding ground-truth views.
  We also show horizontal and vertical EPIs corresponding to red and yellow dotted lines.
  }
  \label{fig:datapairs}
\end{figure} 

\section{Light Field Raindrop Dataset}
% 器材。稳定问题，拍摄的步骤，如何解码。
In this section, we introduce the LFRR dataset construction in details.
Each raindrop removal image pair consists of two LFs with the same background, yet one is degraded by raindrops and the other is raindrop-free.
The Lytro Illum camera is used for the LF acquisition.
In order to eliminate the harmful influence caused by the refraction difference between glasses, different from \cite{qian2018attentive} which uses two glasses for the LF acquisition, we prepare only one piece of transparent glass with 4.5 mm thickness.
We first shoot through the clean glass, then spray water droplets on the glass, and shoot again.
These two shoots are completed with the same camera parameters, such as focal length, ISO, and so on.
In this way, we get a pair of LF raw data.
We set the distance between the glass and the camera lens varying from 1 to 4 cm to generate diverse raindrop images.
To ensure that two LFs have the same background layer, we take measures to alleviate camera motion and scene motion.
For the former, we use a tripod to fix the camera and a shutter cable to control the shooting.
For the latter, we shoot static scenes on a windless and cloudless sunny day.
The LF view images are generated by open source LF toolbox from LF raw data without any post processing, such as color correction and histogram equalization.
% \footnote{\url{https://github.com/doda42/LFToolbox}}

After carefully shooting, decoding and selection, we obtained 32 pairs of high-quality LFRR data, including various background scenes and raindrops.
Some samples are shown in Fig.\ref{fig:datapairs}.
The angular resolution is $7\times 7$, while the spatial resolution is $400 \times 600 $.
Among them, 24 LF pairs constitute the training set, and the remaining 8 LF pairs serve as the validation set.
We will public this dataset once our paper is accepted.

\begin{figure*}[h]
	\centering
	\begin{minipage}{0.99\linewidth}
		\centerline{\includegraphics[width=\linewidth]{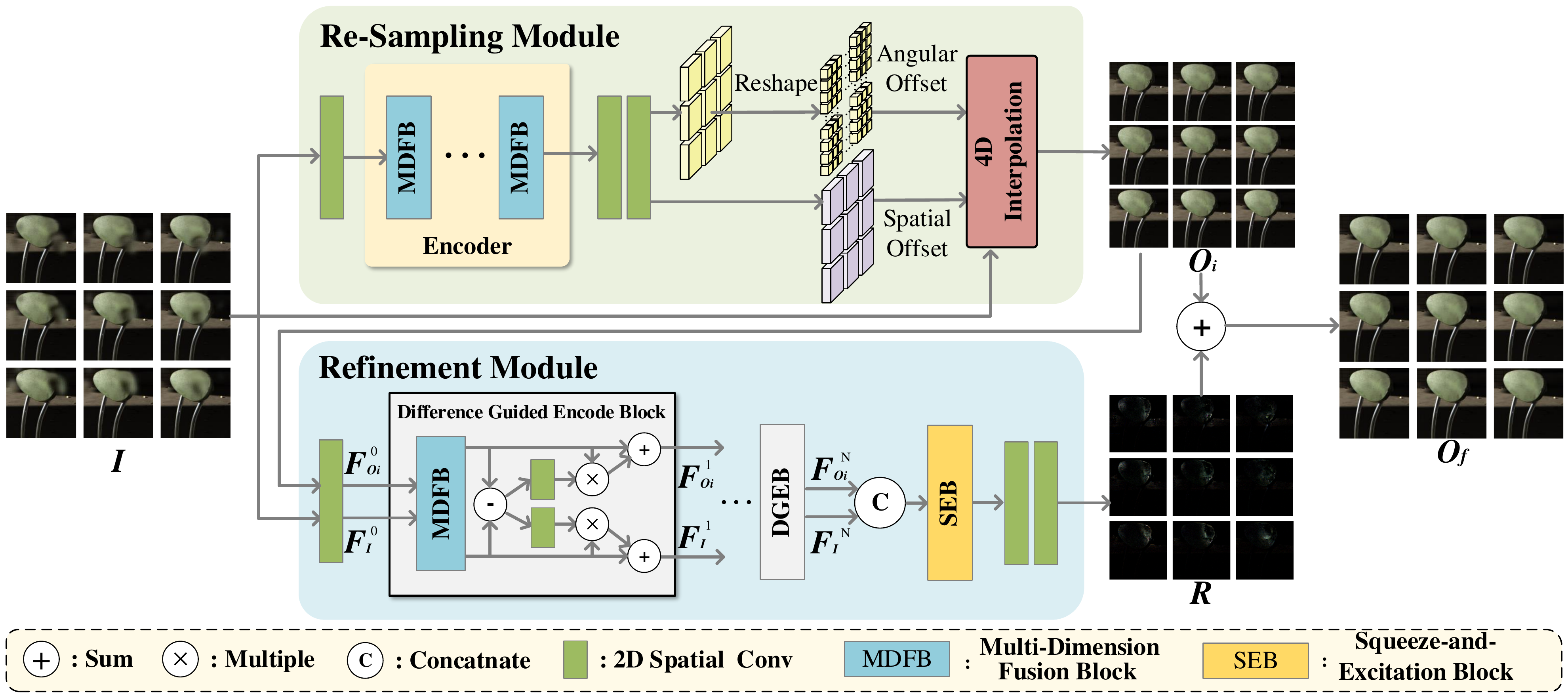}}
	\end{minipage}
	\caption{The overall architecture of the proposed network.}	
	\label{fig:architecture}
\end{figure*}

\section{Methodology}
% In this section, we first introduce the overall architecture of the proposed network, and then introduce the proposed Re-Sampling Module (RSM) and Refinement Module (RM) in details.

\subsection{Overall Architecture}
% 输入的定义，问题的定义，模型的流程
The goal of the LFRR is to restore the background area obscured by raindrops in each view of the LF.
The input LF with raindrops is represented as $I(u,v,x,y) \in \mathbb{R} ^{U\times V \times X\times Y \times 3}$, where $(U,V)$ and $(X,Y)$ are angular and spatial resolutions, respectively, and the number $3$ refers to the RGB color channels.
The overall architecture is divided into two parts: the re-sampling module (RSM) and the refinement module (RM), as shown in Fig.\ref{fig:architecture}.

For each pixel in $I$, the RSM first predicts re-sampling positions, and then carries out the proposed 4D interpolation on $I$ according to re-sampling positions to obtain the sampled LF which is also the initial raindrop removal output $O_i$:
\begin{equation}
    O_i = RSM(I).
\end{equation}

% The RM is designed for two purposes.
% One is to recover the area where is difficult to remove raindrops by 4D sampling, and the other is to correct the pixel error caused by 4D interpolation.
% To better achieve these two objectives, we devise a novel encode block called Difference Guided Encode Block (DGEB) that focus more on regions that need to be refined.
The RM extracts features from $I$ and $O_i$ and predicts a residual map $R$ to refine $O_i$:
\begin{equation}
    R = RM(I, O_i).
\end{equation}
The final raindrop removal output $O_f$ is the summation of $O_i$ and $R$:
\begin{equation}
    O_f = O_i + R,
\end{equation}
where $R$, $O_i$ and $O_f$ have the same size with $I$.

\begin{figure*}[h]
	\centering
	\begin{minipage}{0.99\linewidth}
		\centerline{\includegraphics[width=\linewidth]{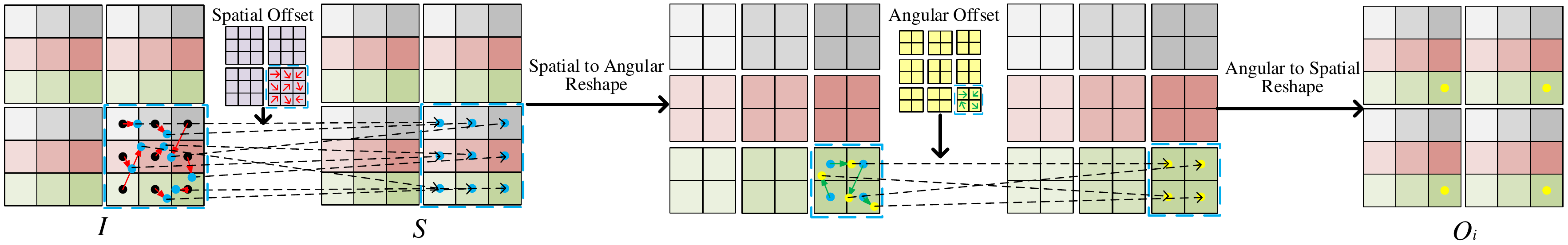}}
	\end{minipage}
	\caption{We take $I$ with the resolution of $(2,2,3,3)$ as instance to illustrate the 4D interpolation.
	The 4D interpolation is realized by first spatial interpolation and then angular interpolation.
	Spatial interpolation is conducted on $I$ to generate $S$, while the angular interpolation works on $S$.
	Since the interpolation kernel is two-dimensional, $S$ and angular offset are reshaped to angular domains before angular interpolation.
	After the angular interpolation, the result is transformed back to the original shape to obtain the 4D interpolation result $O_i$.
    }
	\label{fig:4dInterpolation}
\end{figure*}

\subsection{Re-Sampling Module}
% 出发点，模块的设计。4d插值。
\subsubsection{Motivation}
% Different from rain streak removal task, 
One unique characteristic of the raindrop removal task is that raindrops are closer to the camera and have the shallower depth than the background.
In the LF, the shallower the depth of the object, the larger the disparity it has.
Therefore, in the raindrop LF $I$, most texture details obscured by raindrops in one view are clear in the corresponding areas of other views, as shown in Fig.\ref{fig:intro}.
% This is a great advantage of the LF compared with single image.
% We determine to directly utilize the complementary pixel information in  $I$ by re-sampling for raindrop removal.
% We determine to directly select raindrop-free areas from other views to restore the polluted areas by re-sampling.
Thus, we propose a novel $4D$ re-sampling strategy to select pixels from $I$ and generate one new LF with less raindrop degradation.

% 流程之前聊过了，所以这里直接注释掉。
% The RSM generates the re-sampled LF in three steps, which is shown in Fig. \ref{fig:architecture}.
% The RSM first extracts features from the raindrop LF $I$ by the encoder.
% Then, the RSM predicts the re-sampling position offset for each point in $I$.
% Finally, the 4D interpolation is carried out on $I$ according to the predicted re-sampling positions to generate the re-sampled LF $O_i$.

\subsubsection{Re-sampling Position Prediction}
The first step of the RSM is to calculate the correct re-sampling positions for each point.
Specifically, the re-sampling position $P$ is the summation of the initial position $P_0$ and the re-sampling position offset $\Delta P$ to be predicted.

In the LF, the initial position of each pixel is represented by a vector with length four, including two angular coordinates and two spatial coordinates.
For example, the coordinates of the upper left starting point of the upper left view and its adjacent points on the right are $(0,0,0,0)$ and $(0,0,1,0)$.
Thus, the available $P_0$ is the coordinate set of the whole LF in the size of $\mathbb{R}^{U\times V\times X\times Y \times 4}$.

Then, the core task of this step is to calculate the position offset relative to the initial positions for each point in $I$.
Here, the Multi-Dimension Fusion Block (MDFB) \cite{sun2020multi} is adopted as the feature extraction block, and $M$ MDFBs are cascaded in serial as the encoder.
Next, the offset $\Delta P$ is predicted by two 2D convolutional layers from the features extracted by the encoder:
% The features extracted by the encoder are utilized to predict the sampling offset $\Delta P$ relative to $P_0$ by two 2D convolutional layers:
\begin{equation}
    \Delta P = Conv(Encoder(I)),
\end{equation}
where $\Delta P$ has the same size with $P_0$.
Now, the re-sampling is conducted on the irregular and offset locations $P = P_0 + \Delta P$.
Moreover, we constrain re-sampling positions in $P$ to be located in the range of $I$, hence each re-sampling point is valid.
% Since the interpolation operation is performed on $I$, each re-sampling position in $P$ should be limited into the resolution range of $I$.
% For the position that cross the $I$'s boundary, it is bounded to the nearest boundary position.

\subsubsection{4D Interpolation}
As the offset $\Delta P$ is typically fractional, re-sampling is implemented via interpolation operation.
Since the pixel location in the LF is four-dimensional, the 4D interpolation is necessary.
We separate 4D interpolation into two steps: spatial bilinear interpolation and angular bilinear interpolation, which is shown in Fig.\ref{fig:4dInterpolation}.
The result of 4D interpolation is the initial raindrop removal output $O_i$.
The 4D interpolation is defined as
\begin{equation}
\begin{split}
    S(p) = \sum_{q^I} G_{spa}(p,q^I)\cdot I(q^I), \\
    O_i(p) = \sum_{q^S} G_{ang}(p, q^S) \cdot S(q^S),
\end{split}
\end{equation}
where $p$ denotes a 4D fractional location in $P$, $q^I$ enumerates all integral locations around $p$ in $I$ for spatial bilinear interpolation, $S$ is the spatial bilinear interpolation result from $I$, $q^S$ enumerates all integral locations around $p$ in $S$ for angular bilinear interpolation, $G_{spa}(\cdot, \cdot)$ and $G_{ang}(\cdot, \cdot)$ are spatial and angular bilinear interpolation kernel respectively, $O_i$ is the angular bilinear interpolation result from $S$.
$G_{spa}(\cdot, \cdot)$ and $G_{ang}(\cdot, \cdot)$ are both two-dimensional, and they are separated into two 1D kernels as
\begin{equation}
    \begin{split}
        G_{spa}(p, q^I) = g(q^I_x, p_x) \cdot g(q^I_y, p_y), \\
        G_{ang}(p, q^S) = g(q^S_u, p_u) \cdot g(q^S_v, p_v),
    \end{split}
\end{equation}
where $g(a,b) = max(0, 1-|a-b|)$.
Note that, the order of the spatial interpolation and angular interpolation does not affect the 4D interpolation result.

% It is noted that the difference between the proposed RM and deformable convolution \cite{dai2017deformable} is that the RM make only one position offset prediction for the original input $I$ instead of several sampling position predictions for convolutional kernels.

\subsection{Refinement Module}
\label{refinement}
There are two weaknesses existed in the re-sampling strategy.
On one hand, it is difficult for pixel re-sampling strategy to recover the region completely polluted by raindrops at all views.
% In other words, for some pixels occluded by raindrops, there is no corresponding raindrop-free pixels for reference in $I$.
On the other hand, since the re-sampling position is fractional, the 4D interpolation will inevitably bring slight pixel errors, e.g., pixels in clean backgrounds are slightly affected by surrounding pixels.
% The RM is designed to address these two shortcomings.

The RM is a module strongly coupled with the RSM for addressing shortcomings of the RSM.
As shown in the bottom part of Fig.\ref{fig:architecture}, the RM first extracts features from $I$ and $O_i$, and then generates a residual map $R$ to refine $O_i$.
The raindrop LF $I$ is introduced to the RM for ensuring information integrity.

Specifically, $I$ and $O_i$ are first separately fed to a weight-shared 2D convolution layer which works on spatial dimensions to generate the initial feature $F^{0}_I$ and $F^{0}_{O_i}$.
Considering that the difference between $I$ and $O_i$ reflects on the areas of raindrops and wrong re-sampled backgrounds, the difference is beneficial for identifying the areas to be refined.
Therefore, we propose a novel Difference Guided Encode Block (DGEB) to focus more on areas where need improvement by leveraging the guidance of the difference between $I$ and $O_i$.

As shown in the gray box of the Fig.\ref{fig:architecture}, the first DGEB takes the pair of $F^{0}_I$ and $F^{0}_{O_i}$ as inputs to achieve interaction.
Firstly, $F^{0}_I$ and $F^{0}_{O_i}$ are separately encoded by a weight-shared MDFB, and the generated features are called the encoded $F^{0}_I$ and the encoded $F^{0}_{O_i}$.
Then, the difference feature $F^{1}_{diff}$ are calculated by the element-wise subtraction between the encoded $F^{0}_I$ and the encoded $F^{0}_{O_i}$.
Next, the difference feature $F^{1}_{diff}$ are fed to two $1\times1$ convolution layers to produce two single-channel attention maps.
Finally, these two attention maps are separately fused with the encoded $F^{0}_I$ and the encoded $F^{0}_{O_i}$ by element-wise multiplication and summation.
The outputs of the first DGEB are denoted as $F^{1}_I$ and $F^{1}_{O_i}$.
In RM, we cascade $N$ DGEBs for feature extraction and interaction, i.e., the outputs of a DGEB form the inputs of its subsequent DGEB, and the final output features are represented as $F^{N}_I$ and $F^{N}_{O_i}$.
In summary, the DGEB can be formulated as
\begin{equation}
    \begin{split}
    F^k_{diff} = H^{(k)}(F^{k-1}_{I}) - H^{(k)}(F^{k-1}_{O_i}), \\
    F^{k}_{I} = H^{(k)}(F^{k-1}_{I}) \times Conv^{(I,k)}_{1\times1}(F^k_{diff}) + H^{(k)}(F^{k-1}_{I}), \\
    F^{k}_{O_i} = H^{(k)}(F^{k-1}_{O_i}) \times Conv^{(O_i,k)}_{1\times1}(F^k_{diff}) + H^{(k)}(F^{k-1}_{O_i}), \\
    (k=1,2,\cdots,N)
\end{split}
\end{equation}
where $F^{k}_{I}$ and $F^{k}_{O_i}$ represent the output features corresponding to $I$ and $O_i$ of the $k^{th}$ DGEB, respectively, $H^{(k)}$ represent the MDEB in the $k^{th}$ DGEB.

The outputs $F^{N}_I$ and $F^{N}_{O_i}$ are then concatenated along the channel dimension.
Inspired by \cite{hu2018squeeze}, the Squeeze-and-Excitation Block (SEB) is adopted here for better information fusion.
Suppose the input feature is $F$, The calculation process of the SEB is:
\begin{equation}
    SEB(F) = F \times \sigma (Linear(Linear(AvgPool(F)))),
\end{equation}
where $AvgPool$ is the global spatial average pooling, $Linear$ refers to a full-connection layer, and $\sigma$ is the sigmoid function.
The residual map $R$ is produced by 2D spatial convolution layers from the fused features.
The final raindrop removal output $O_f$ is the summation of $O_i$ and $R$.

% Taking $I$ and $O_i$ as inputs has two advantages.
% The first is to ensure the integrity of information.
% Specifically, 4D interpolation re-sampling may brings some improper re-sampled pixels and breaks local structural characteristics which is hard to supplement without the original input $I$.
% Specifically, 4D interpolation re-sampling may bring wrong re-sampled pixels and destroys local structural characteristics, and losses some effective information which is hard to supplement without the original input $I$.
% Second, the difference region between $I$ and $O_i$ is more concentrates on raindrop areas and mistaken re-sampling pixels which are also areas to be refined.
% Therefore, we propose a novel Difference Guided Encode Block (DGEB) based on MDFB to focus more on areas where need improvement by leveraging the guidance of the difference between $I$ and $O_i$, as shown in the middle of the Fig. \ref{fig:architecture}.

% Suppose the input features of the DGEB are $F_I$ and $F_{O_i}$, corresponding to features extracted from $I$ and $O_i$ respectively.
% In DGEB, they are first encoded by a weight-shared MDFB.
% The difference feature $F_d$ is the difference between the encoded $F_{O_i}$ and the encoded $F_I$.
% Then, two 2D spatial convolution layer with output channel of 1 are applied to generate two spatial attention map from $F_d$ for the encoded $F_I$ and the encoded $F_{O_i}$.
% The attention map and corresponding feature are fused by element-wise multiplication and summation.

\subsection{Learning Objective}
% 损失函数
% 训练过程中训练目标的改变？
Besides the widely used Mean Absolute Error (MAE), we utilize the SSIM function \cite{wang2004image} and EPI gradient loss \cite{jin2020learning} as additional evaluation metrics for the generated LF.
These three loss functions are denoted as $L_{MAE}$, $L_{SSIM}$ and $L_{EPI}$.
The combination of the MAE loss function and SSIM loss function can preserve the per-pixel similarity as well as the local structure.
The EPI gradient loss is beneficial for enhancing view consistency of the output LF.

Given the raindrop-free ground truth $Y$ and the model's output $\hat{Y}$, the learning Objective of $\hat{Y}$ is defined as:
\begin{equation}
    L(\hat{Y}, Y) = L_{MAE}(\hat{Y}, Y) + \alpha L_{SSIM}(\hat{Y}, Y) + \beta L_{EPI}(\hat{Y}, Y),
\end{equation}
where $\alpha$ and $\beta$ denote the hyper parameters used to balance losses, which are set to 0.1 and 1 in the experiment.

The whole training process of the proposed model is divided into two stage.

In the first stage, both $O_i$ and $O_f$ are supervised by the raindrop-free ground truth $Y$.
The supervision of $O_i$ is necessary during the initial training, because it enables the model to effectively search the raindrop-free backgrounds in $I$.
The learning objective of this stage is
\begin{equation}
\label{l_s1}
    L_{stage1} = L(O_f, Y) + \lambda L(O_i, Y),
\end{equation}
where the weight hyper parameter $\lambda$ is set to 0.5 in the experiment.
When the model converges, the training enters the second stage.

In the second stage, we remove the supervision of $O_i$ to make the model pay attention to learning the prediction of the final result $O_f$.
The learning objective of the second stage is
\begin{equation}
\label{l_s2}
    L_{stage2} = L(O_f, Y).
\end{equation}
When the model converges again, the training ends.
The second training stage brings slight performance improvement (about 0.15dB PSNR).

\section{Experiment}
\subsection{Comparison Methods}
We compare the proposed model with six 2D image rain removal methods: AttGAN \cite{qian2018attentive}, RSECAN \cite{li2018recurrent}, SDA \cite{quan2019deep}, PIDN \cite{ren2019progressive}, DualRes \cite{liu2019dual} and MSPIR \cite{zamir2021multi}. 
Among them, RSECAN, PIDN and MSPIR are rain streak removal algorithms.
Because these three algorithms do not model the physical properties of rain streaks, but make improvements in the design of the convolutional networks, all of them can be transferred to the raindrop removal task.

For a more convincing comparison, four LF spatial super resolution methods are adjusted for LFRR task which are SAS \cite{yeung2018light}, MDFN \cite{sun2020multi}, InterNet \cite{wang2020spatial}, MEGNet \cite{zhang2021end}.
As mentioned in Section \ref{LFP}, algorithms in LF spatial super resolution domain reconstruct all views of the LF by extracting complementary information between different views, which has many similarities to solve the LFRR problem.
Besides, these four methods are convenient to be adjusted to handle LFRR problem by removing the up-sampling operations.

\subsection{Implementation Details}

We train and validate models on the proposed LFRR dataset.
For 2D methods, views in LFs are separated, e.g., a pair of LFs with $7 \times 7$ angular resolutions is equivalent to $49$ 2D image pairs.
We augment training data for 2D methods by random cropping, flipping, rotating and resizing.
The popular $4$ LF super resolution methods are carefully adjusted by removing their up-sampling operations.
Apart from selecting appropriate initial learning rate, we maintain the original configuration of algorithms involved in comparisons.
The quantitative evaluation is done in terms of two metrics: PSNR and SSIM calculated on the luminance (Y channel of the YCbCr space).

The proposed model is implemented by PyTorch.
All experiments are run on one NVIDIA GeForce GTX A4000.
The designed model has $M=5$ MDFBs and $N=5$ DGEBs.
Each MDFB or DGEB has the same number of input and output channel $c=64$.
In training, we keep the angular size $7 \times 7$ and randomly crop patches of spatial size $96 \times 96$ as inputs.
To relieve the over-fitting problem on the premise of protecting the structural characteristics of LFs, we augment the training set by random global flipping and rotating for 4D approaches.
The proposed model is trained by Adam \cite{kingma2014adam} optimizer with batch size of 1.
The initial learning rate is $3e^{-4}$, and the number of epochs is set to $4000$.
We use cosine annealing strategy to adjust learning rate during the training, and set the final minimum learning rate to 0.
% The initial learning objective is Equation \ref{l_s1}.
% When the model converges for the first time, the learning objective becomes Equation \ref{l_s2}.
The total training time for the proposed model is approximately 2 days.

% 要说清楚4d方法的调整细节。
% 几个方面的比较：1.指标 2.可视化（包含视角一致性） 3.参数量 4.2d方法表现不佳的原因：2d方法难以获取被雨滴影响的背景区域的信息，只能够通过在训练集上学习到的模式来做恢复，这部分的工作更像是inpainting，且需要大规模数据集的支撑，在小样本上表现不佳。相比之下，基于光场的方法更关注对输入信息的提取，对数据集规模的依赖小很多。

\begin{figure*}[h]
	\centering
	\begin{minipage}{.99\linewidth}
		\centerline{\includegraphics[width=.99\linewidth]{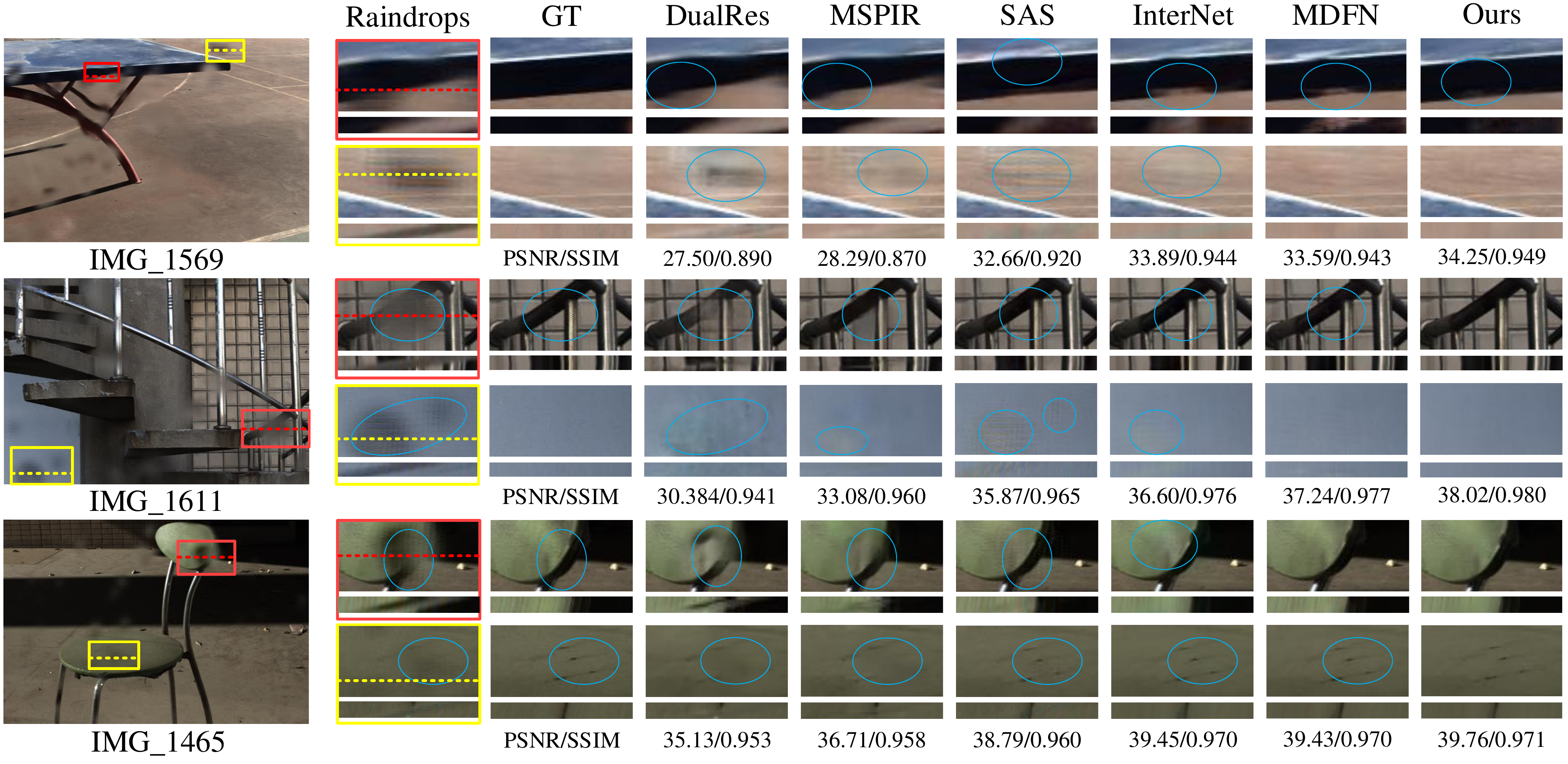}}
	\end{minipage}
	\caption{Qualitative comparisons of top raking approaches. 
	Two local areas of central views in the red and yellow boxes are enlarged for comparison.
	In addition, we also visualize the horizontal EPIs corresponding to the dotted lines to show the view consistency.
	By contrast, our results show clear details in restored backgrounds, and our EPIs also show consistent lines compared with ground truth.
	}
	\label{fig:comparisons}
\end{figure*}

\subsection{Qualitative Evaluation}
See Fig. \ref{fig:comparisons} for the visualization of some results of five top ranking methods.

In image restoration quality, 2D methods generate significant artifacts and wrong textures, and even poorly affect raindrop-free backgrounds, which can be attributed to the limited scale of training set and the lack of details in the occluded background.
For other 4D methods, the restored background areas are not smooth and natural enough.
These methods rely on convolutional neural network to implicitly predict the residual between input and output, which is unstable.
By comparisons, our model first removes raindrops by directly utilizing useful pixel information in raindrop LFs, which reduces the dependence on and the difficulty of residual learning.
In Fig.\ref{fig:comparisons}, our results show the clearest and most real details in the restored backgrounds.

Because the LFRR is to deal with all views of the LF, the view consistency is also an important measurement.
In Fig.\ref{fig:comparisons}, we show horizontal EPIs corresponding to the dotted lines for view consistency comparisons.
While the EPIs of other methods contain noisy artifacts, our EPIs have the most consistent lines compared with ground truth.

\begin{table*}
\centering
\caption{Quantitative results of different methods on the proposed dataset.}
\renewcommand\arraystretch{1.2}
\scalebox{1}{

\resizebox{\linewidth}{!}{
  \begin{tabular}{c|c |c|c | c| c| c|c | c | c | c | c}
        \toprule[1pt]
        \multirow{3}{*}{\textbf{Metrics}} &
        \multicolumn{6}{c|}{\textbf{2D Methods}} & \multicolumn{5}{c}{\textbf{4D Methods}} \\
        % \cmidrule(r){3-8}\cmidrule(r){9-14}
        \cline{2-6}\cline{7-12}
        \multicolumn{1}{c|}{}&
        \makecell[c]{AttGAN\\(CVPR18)} & 
        \makecell[c]{RSECAN\\(ECCV18)} & 
        \makecell[c]{SDA\\(ICCV19)} & 
        \makecell[c]{PIDN\\(CVPR19)} & 
        \makecell[c]{DualRes\\(CVPR19)} & 
        \makecell[c]{MSPIR\\(CVPR21)} & 
        \makecell[c]{SAS\\(TIP18)} & 
        \makecell[c]{MDFN\\(Arxiv20)} & \makecell[c]{InterNet\\(ECCV20)} & 
        \makecell[c]{MEGNet\\(TIP21)} & \textbf{Ours}  \\
        \hline
        {\textbf{PSNR}}&30.3670 & 31.8507 & 27
        0175 & 31.7731 & 31.8626 & 33.0273 & 36.2761 & 37.4302 & \underline{\textit{37.6900}} & 37.1023 & \textbf{37.9671}  \\
        {\textbf{SSIM}}& 0.9341 & 0.9393 & 0.9098 & 0.9317 & 0.9431 & 0.9440 & 0.9575 & 0.9710 & \underline{\textit{0.9716}} & 0.9681 & \textbf{0.9734} \\ 
        {\textbf{Params}}&  6.2M & 0.26M & 7.2M & 0.17M & 10M & 3.6M & 0.74M & 0.48M & 8M & 0.42M & 0.4M \\ 
    %   \cline{2-14}
    \bottomrule
    \end{tabular}
    }
}
\label{tab:comparisons}
\end{table*}
\subsection{Quantitative Evaluation}
The quantitative comparisons are shown in Table.\ref{tab:comparisons}.
By comparison, our model achieves the best performance in both PSNR and SSIM with light-weight parameters.
It is noted that compared with the second best method InterNet, our model generates better results with only one twentieth of its learnable parameters. 

According to the Table.\ref{tab:comparisons}, the overall performance of 2D methods is not as good as that of 4D methods.
2D models benefit for more available data sets and higher upper limitation of model parameters than 4D models.
However, when training in the current small-scale training set, these 2D methods have two obvious disadvantages.
Firstly, because texture details covered by raindrops in single image are completely lost, when restoring the occluded background areas, 2D methods rely more on the inpainting patterns learned from the training data, while the small-scale training set provides little inpainting patterns.
Secondly, compared with the LF based methods, due to the lack of scene structure characteristics in single image, 2D methods are weak in the design of the learning objective.

It can be seen that SDA performs worse than other 2D approaches.
The method SDA introduces the edge map as additional input, and builds module to exploit the physical shape properties of raindrops, including closedness and roundness.
However, since the camera focuses on the background when shooting, raindrop areas in our dataset have severe blur and grid effects, which brings great trouble to SDA.
Besides, some scenes in our dataset contain rain stripes that are also hard for SDA to deal with.

Compared with 2D inputs, the LF with unique structural characteristics provides abundant complementary information, which enables 4D approaches more dependent on the input LF rather than the inpainting patterns learned from the training set.
Therefore, 4D methods can achieve great raindrop removal performance by training on a small-scale dataset.

\setlength{\tabcolsep}{2.5mm}{
\begin{table}[]
\centering
\caption{Ablation study.}
\renewcommand\arraystretch{1.1}
\scalebox{0.7}{
\resizebox{\linewidth}{!}{
  \begin{tabular}{c c|c c|c c }
\toprule[1pt]
\multicolumn{2}{c|}{\textbf{Interpolation}} & \multicolumn{2}{c|}{\textbf{Refinement}} &
\multicolumn{2}{c}{\textbf{Metrics}}\\
% \cmidrule(r){3-8}\cmidrule(r){9-14}
\cline{1-6}
Spa & Ang  &MDFB & DGEB & PSNR & SSIM \\
\hline
\checkmark & & & \checkmark & 36.7829& 0.9655\\
 & \checkmark &  & \checkmark & 37.8123 & 0.9725\\
\checkmark & \checkmark &  &  & 36.6814 & 0.9679 \\
\checkmark &  \checkmark &  \checkmark&  & 37.9210 &  0.9725\\
\checkmark & \checkmark & & \checkmark & 37.9671 & 0.9734 \\
\bottomrule
\end{tabular}
}      
}
\label{tab:ablation}
\end{table}
}

\section{Ablation Study}
% To begin with, we construct the baseline model by only retaining the MDFB based encoder for residual prediction.
% The encoder in the baseline model has $M=10$ MDFBs with the same input and output channel $c=80$ to ensure similar parameter quantities, and its validation result is shown in the first row of Table.\ref{tab:ablation}.

\subsection{Effects of Re-sampling Patterns}
\label{ablationResampling}
Apart from the 4D re-sampling in both spatial and angular dimensions mentioned above, re-sampling can be separately conducted in the spatial or angular dimensions.
Since the re-sampling module and the refinement module are strongly coupled, we conduct two ablation experiments with the refinement module to evaluate re-sampling patterns:
\begin{itemize}
    \item 2D spatial re-sampling with the refinement module.
    \item 2D angular re-sampling with the refinement module.
\end{itemize}

In both cases, the model predicts 2D location offset, then performs the bilinear interpolation on $I$ in corresponding dimensions according to the predicted re-sampling positions. 
The results are listed in the first two lines of Table~\ref{tab:ablation}.

It is hard to recover the occluded background areas by spatial re-sampling due to the lack of texture details.
The useless spatial re-sampling also hinders the subsequent processing.
As shown in the first row of Table.\ref{tab:ablation}, the performance of the model using spatial re-sampling is the worst.

When re-sampling only on angular dimensions, the model can obtain effective information from other views to supplement background details.
However, due to the disparity and occlusion existed in backgrounds, angular re-sampling strategy is difficult to accurately obtain the details of the occlusion boundaries in backgrounds.
By comparison, 4D re-sampling can alleviate this problem by providing a wider search space.
In the second row of Table.\ref{tab:ablation}, the model using angular re-sampling performs better than the spatial re-sampling model and worse than the 4D re-sampling model, which not only verifies the effectiveness of the angular re-sampling, but also confirms the superiority of 4D re-sampling.
It is worth mentioning that the proposed model using angular re-sampling already performs better than all approaches involved in comparisons.

\subsection{Effects of the Refinement Module}
As mentioned in Section~\ref{refinement}, the RM is designed to solve the shortcomings of the re-sampling strategy.
In this subsection, we conduct two ablation experiments to separately verify the effectiveness of the RM and the DGEB:
\begin{itemize}
    \item Abandon the refinement module.
    \item Replace the DGEB in the refinement module with the MDFB.
\end{itemize}

For the former ablation experiment, we enlarge the channel number of the RSM to $80$ to ensure the similar parameter quantities compared with the complete model.
The corresponding results are listed in the third and forth lines of the Table~\ref{tab:ablation}.
Without the RM, the model's performance decreases a lot, e.g., the PSNR decline of $1.29$dB is observed, which illustrates that the RM is essential and significant.
The comparison between the forth line and the fifth line show that the DGEB brings slight PSNR and SSIM improvements.
% Note that, because the MDFB based model (the forth line) has achieved great performance, the improvement brought by the DGEB are hard won.
% DGEBs in the RM are replaced by MDFBs to validate the effectiveness of the DGEB.
% As shown in the fourth row of Table.\ref{tab:ablation}, without DGEB, the model's performance decreases a lot in both PSNR and SSIM.
% This demonstrates that the DGEB can benefit the refinement.

\begin{figure*}[h]
	\centering
	\begin{minipage}{\linewidth}
		\centerline{\includegraphics[width=.99\linewidth]{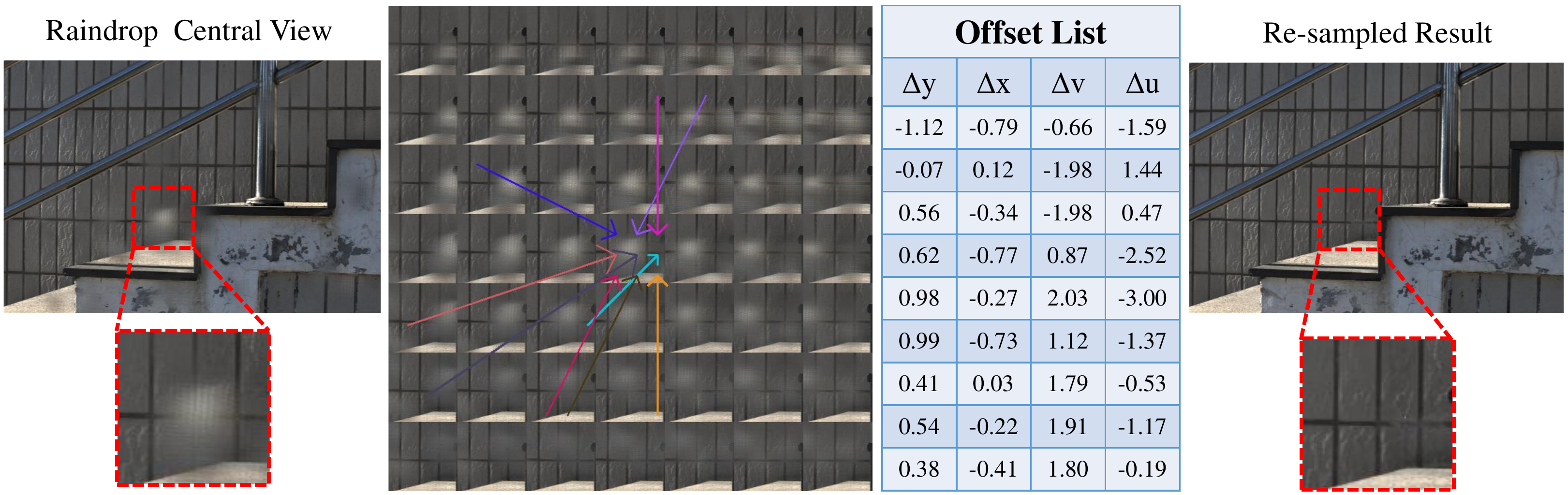}}
	\end{minipage}
	\caption{
	A visual example of the 4D re-sampling procedure. 
    The re-sampling processes of nine points in a raindrop area of the central view are visualized.
	For visualization, the re-sampling coordinates in decimal form are rounded to integers.
	In offset list, we show the re-sampling offset of these nine points in order.
	}
	\label{fig:resampling}
\end{figure*}

\begin{figure*}[h]
	\centering
	\begin{minipage}{\linewidth}
		\centerline{\includegraphics[width=.99\linewidth]{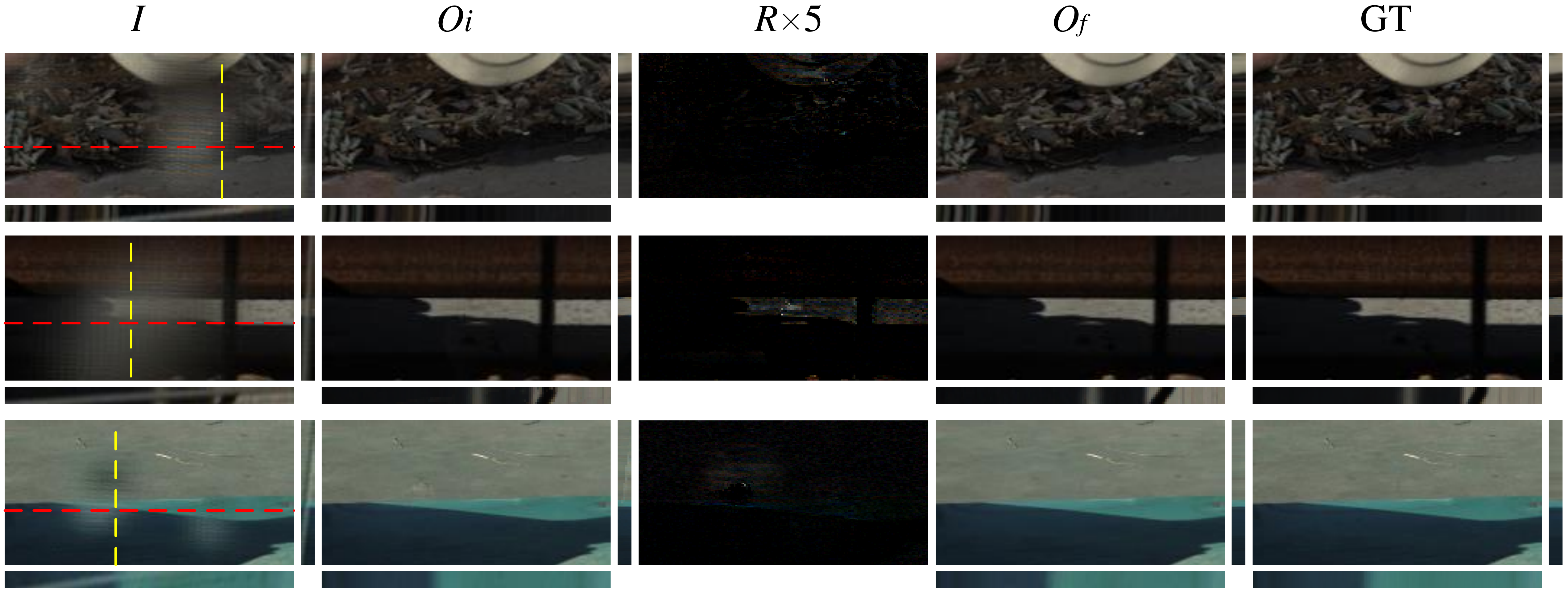}}
	\end{minipage}
	\caption{Outputs corresponding to central views produced by each module in our network.
	$R$ is amplified for visualization.
	}
	\label{fig:results}
\end{figure*}

\subsection{Performance Analysis}
In this subsection, we discuss more about the performance of each module.

In Fig.~\ref{fig:resampling}, we show a visual instance of the 4D re-sampling procedure.
The 4D re-sampling procedure illustrates that points of raindrop areas can re-sample the pixels with less raindrop pollution in other views.
According to the offset list, 4D re-samplings in raindrop area are dominated by the offset in angular domains, while the spatial offset expands the sampling range on the basis of angular offset, which is accord with our analysis about the spatial and angular re-sampling in Section \ref{ablationResampling}.
The re-sampled result show that the re-sampling module directly utilizes the complementary pixel information between different views to produce pretty initial raindrop removal result.

In Fig.\ref{fig:results}, more intermediate results produced by our model are visualized.
% We can see that the proposed re-sampling operation can effectively obtain the complementary pixel information from $I$, so as to restore the obscured areas.
We can see that the residual map $R$ generated by the refinement module plays an auxiliary role to make $O_i$ more  natural by providing some high-frequency information, which illustrates that the 4D re-sampling strategy reduces the dependence on and the difficulty of implicit residual learning.

\section{Conclusions}
In this paper, we illustrate the advantages of the light field in raindrop removal task compared with single image.
We propose a novel light field raindrop removal approach which consists of the re-sampling module and the refinement module.
The re-sampling module generates a novel light field with less raindrop pollution through position prediction and 4D interpolation, while the refinement module improves the re-sampled light field.
Furthermore, we build the first real scene light field raindrop removal dataset.
Experiments demonstrate the effectiveness of our model.

% \clearpage\mbox{}Page \thepage\ of the manuscript.
% \clearpage\mbox{}Page \thepage\ of the manuscript.

% This is the last page of the manuscript.
% Now we have reached the maximum size of the ECCV 2022 submission (excluding references).
% References should start immediately after the main text, but can continue on p.15 if needed.

\clearpage
\bibliographystyle{splncs04}
\bibliography{egbib}
\end{document}